\documentclass[conference]{IEEEtran}
 
\def\BibTeX{{\rm B\kern-.05em{\sc i\kern-.025em b}\kern-.08emT\kern-.1667em\lower.7ex\hbox{E}\kern-.125emX}}
    
\usepackage{graphicx} 
\usepackage{makecell}
\usepackage{listings}
\usepackage{multirow}
\usepackage{pgfplots}
\pgfplotsset{compat=1.18}
\usepackage{tikz}
\usepackage{url}
\usepackage{hyperref}
\usepackage{amsmath}

\newtheorem{Definition}{Definition}
\def\checkmark{\tikz\fill[scale=0.4](0,.35) -- (.25,0) -- (1,.7) -- (.25,.15) -- cycle;}

\colorlet{punct}{red!60!black}
\definecolor{background}{HTML}{EEEEEE}
\definecolor{delim}{RGB}{20,105,176}
\colorlet{numb}{magenta!60!black}

\lstdefinelanguage{json}{
    basicstyle=\normalfont\ttfamily,
    numbers=left,
    numberstyle=\scriptsize,
    stepnumber=1,
    numbersep=8pt,
    showstringspaces=false,
    breaklines=true,
    frame=lines,
    backgroundcolor=\color{background},
    literate=
     *{0}{{{\color{numb}0}}}{1}
      {1}{{{\color{numb}1}}}{1}
      {2}{{{\color{numb}2}}}{1}
      {3}{{{\color{numb}3}}}{1}
      {4}{{{\color{numb}4}}}{1}
      {5}{{{\color{numb}5}}}{1}
      {6}{{{\color{numb}6}}}{1}
      {7}{{{\color{numb}7}}}{1}
      {8}{{{\color{numb}8}}}{1}
      {9}{{{\color{numb}9}}}{1}
      {:}{{{\color{punct}{:}}}}{1}
      {,}{{{\color{punct}{,}}}}{1}
      {\{}{{{\color{delim}{\{}}}}{1}
      {\}}{{{\color{delim}{\}}}}}{1}
      {[}{{{\color{delim}{[}}}}{1}
      {]}{{{\color{delim}{]}}}}{1},
}

\begin{document}

\newcommand{\ls}[1]{{\color{red}{LS: #1}}}

\title{Extending the SAREF4ENER Ontology with Flexibility Based on FlexOffers
}
\author{\IEEEauthorblockN{Fabio\\Lilliu}
\IEEEauthorblockA{
University of Cagliari\\
fabio.lilliu86@unica.it}
\and
\IEEEauthorblockN{Amir\\Laadhar}
\IEEEauthorblockA{
PANTOPIX GmbH \& Co. KG\\
amir.laadhar@pantopix.com}
\and
\IEEEauthorblockN{Christian\\Thomsen}
\IEEEauthorblockA{
Aalborg University\\
chr@cs.aau.dk}
\and
\IEEEauthorblockN{Diego \\Reforgiato Recupero}
\IEEEauthorblockA{
University of Cagliari\\
diego.reforgiato@unica.it}
\and
\IEEEauthorblockN{Torben Bach\\Pedersen}
\IEEEauthorblockA{
Aalborg University\\
tbp@cs.aau.dk}
}

\IEEEoverridecommandlockouts
\maketitle
\begin{abstract}
A key element to support the increased amounts of renewable energy in the energy system is flexibility, i.e., the possibility of changing energy loads in time and amount. Many flexibility models have been designed; however, exact models fail to scale for long time horizons or many devices. Because of this, the FlexOffers model has been designed, to provide device-independent approximations of flexibility with good accuracy, and much better scaling for long time horizons and many devices. An important aspect of the real-life implementation of energy flexibility is enabling flexible data exchange with many smart energy appliances and market systems, e.g., in smart buildings. For this, ontologies standardizing data formats are required. However, the current industry standard ontology for integrating smart devices for energy purposes, SAREF for Energy Flexibility (SAREF4ENER), only has limited support for flexibility and thus cannot support important use cases. In this paper, we propose an extension of SAREF4ENER that integrates full support for the complete FlexOffer model, including advanced use cases, while maintaining backward compatibility. This novel ontology module can accurately describe flexibility for advanced devices such as electric vehicles, batteries, and heat pumps. It can also capture the inherent uncertainty associated with many flexible load types. 

\end{abstract}
\maketitle

\section{Introduction}
The last decades have seen an increasing usage of renewable energy, and the development of new technologies and concepts to enable its exploitation. In particular, among those new concepts, one has crucial importance for handling the intermittency that comes with renewable energy generation: \emph{flexibility}~\cite{LI2021100054}, i.e., the capability of changing energy loads in time or amount of energy. 
Flexibility enables optimizing energy consumption to achieve several objectives: renewable energy usage maximization~\cite{scheller}, peak shaving~\cite{FOTEINAKI2020109804,CHEN2021110932}, CO2 emissions minimization~\cite{HARRIS20141}, energy cost minimization~\cite{Mugnini2021,MAROTTA2021102392}, provision of ancillary services~\cite{saele}, congestion prevention~\cite{Fonteijn}, and demand response~\cite{CHEN2018125}.
Several models have been developed for flexibility. Many of them are accurate in their description of flexibility, but scale poorly with respect to the length of time horizons and the number of devices. A model that solves these issues is FlexOffers~\cite{DBLP:conf/smartgridcomm/PedersenSN18}, a device-independent model that represents flexibility by using energy constraints that approximate the amount of available flexibility, but scales very well in scenarios with many devices or long time horizons. FlexOffers generates flexoffers (FOs) objects, which describe flexibility from the considered devices.
FlexOffers is a powerful model used in more than 30 European and national projects, whose real-life implementation has been advancing in the last decade. An important aspect is its interaction with smart devices/buildings and the ontologies representing them. To date, a widely used ontology for smart buildings is the Smart Application REFerence (SAREF)~\cite{SAREF} ontology. Specifically, several extensions of SAREF have been created for different domains: SAREF for Energy (SAREF4ENER)~\cite{SAREF4ENER} has been designed to represent energy devices and services, and is the de facto standard for that. However, SAREF4ENER fails to capture energy flexibility for many use cases, specifically when flexibility constraints become more complex: this is the case for advanced devices such as batteries, electric vehicles (EVs), and heat pumps. In addition, there are only few other ontologies representing energy flexibility. One of them is EFOnt~\cite{Comparison1}, a semantic ontology focused on energy flexibility at the building level and its applications. Another is the MAS$^2$TERING ontology~\cite{Comparison2}, which aims to integrate the Universal Smart Energy Framework (USEF) standards with the context of multi-agent systems to support the interoperability of smart grid management agents. Finally, the BIPV-BES-BEF ontology~\cite{Comparison3} targets photovoltaic, battery, and building operation and maintenance, and includes a class that describes flexible control strategies associated with buildings. However, none of these ontologies describe the flexibility constraints of the different types of devices and, like SAREF4ENER, they fail to properly represent flexibility in the aforementioned use cases. This raises the need for an ontology capable of handling these cases.  
To meet this challenge, an ontology called dCO~\footnote{\url{https://w3id.org/dco}}~\cite{DBLP:conf/esws/LaadharTP22} (d is the first letter in the EU project name, removed for double-blind peer review, and CO stands for Common Ontology) 
has been proposed to ensure full flexibility support in the interoperability of smart buildings and services. The contribution we bring in this paper is the dCO flexibility module, which extends SAREF4ENER to better represent all aspects of energy flexibility and enable more powerful flexibility services. It enables compatibility with the full Flexoffers model and constraints, ensuring that the most relevant flexible devices and use cases for them are supported. This is valuable because the resulting ontology, dCO, allows developers to create applications that offer: i) a more accurate representation of flexibility compared to the current state of the art, ii) applicability to a wider range of devices, such as batteries, EVs, and heat pumps, and iii) support for the fast load optimization and (dis)aggregation provided by FlexOffers.

In this paper, we show how the flexibility module of dCO extends SAREF4ENER, how it represents the full Flexoffer model and constraints, and some advanced use cases that can be represented by it.
The remainder of this paper is organized as follows. Section~\ref{section:preliminaries} describes FOs and the SAREF4ENER and dCO ontologies, while 
Section~\ref{section:extension} shows how dCO extends SAREF4ENER and enables full FO representation.  Section~\ref{section:evaluation} evaluates the dCO flexibility module, showing which use cases can and cannot be represented by FOs and SAREF4ENER. Finally, Section~\ref{section:conclusion} concludes the work and provides ideas for future research. 

\section{Preliminaries}
\label{section:preliminaries}

\subsection{Running Example}
We start by making a running example so that we can describe the Flexoffer concept in the following subsections. 
Here, we will consider the case of a heat pump operating in a room. The heat pump is a \textit{Daikin Altherma}
~\cite{DBLP:conf/eenergy/LilliuPS23}, with maximum operational power $P_{max} = 4.6$ kW and coefficient of performance (COP) $3.65$. This model supports the SG-Ready interface~\cite{DBLP:conf/isgteurope/FischerTS18,DBLP:conf/eenergy/BrusokasPSZC21}, a standardized smart grid communication protocol that enables seamless integration with modern energy management systems.
The room has a rectangular section with sides of $4$ and $5$ meters respectively, and it is $3$ meters high. The heat transfer coefficient is $c_{ht} = 6\frac{W}{m^2K}$~\cite{JAYAMAHA1996399}, and heat is dispersed through one of the $3$m x $4$m side walls. The room temperature is $295$K ($22^\circ$C), while the outside temperature is $275$K ($2^\circ$C). We assume that room temperature must stay between $293$K ($20^\circ$C) and $297$K ($24^\circ$C), respectively, because of user settings. 
For reference, the amount of energy (heat) needed to maintain the temperature to $295$K for one hour can be obtained by
\begin{equation*}
    H = (3\cdot4)m^2 \cdot 6\frac{W}{m^2K} \cdot (295K - 275K) \cdot 3600s = 1.44 kWh.
\end{equation*}
Since $COP = 3.65$, the amount of electrical energy needed to provide this amount of heat is
$E = \frac{H}{COP}  = 0.395 kWh.$

\subsection{FlexOffer Life Cycle}

The Flexoffers model describes flexibility in a unified format for different device types~\cite{DBLP:conf/smartgridcomm/PedersenSN18,Lilliu2021}, by providing an approximated representation of flexibility. In exchange for the small amount of accuracy lost with respect to exact models, modeling flexibility with FOs allows for much better scalability over long time horizons and high numbers of devices. FOs provide constraints for energy consumption: \emph{optimizing} an FO means solving an optimization problem on energy consumption (e.g., cost minimization) within the constraints given by the FO. The solution of this optimization problem is a set of values $(e_1, \ldots, e_T)$ for energy consumption at each time unit: this is called a \emph{schedule}. FOs can also be \emph{aggregated}: this means that, from $M$ FOs, it is possible to generate $N \ll M$ FOs that represent their combined flexibility, with some losses. The opposite process is called \emph{disaggregation}: from a schedule for an aggregated FOs, it is possible to generate schedules for the original FOs so that their combined energy consumption is the same as the original schedule. 

\begin{figure}
    \centering
    \includegraphics[scale=.42]{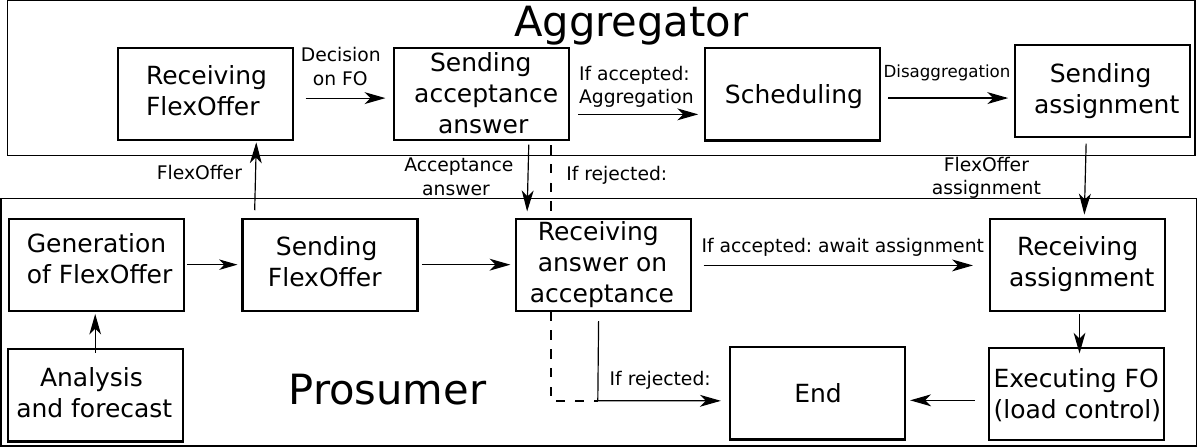}
    \caption{A schematic description of the FOs life cycle (taken from~\cite{DBLP:conf/eenergy/LilliuPSN23}).}
    \label{fig:FOLifeCycle}
\end{figure}

The FO life cycle is shown in Figure~\ref{fig:FOLifeCycle}. Two main actors are involved: an automatic agent representing the prosumer/device owner, which creates and executes the FOs, and another one who manages and optimizes the FOs, which is often either the Balance Responsible Party (BRP) or the aggregator: for clarity, in this paper we will assume it to be the aggregator~\cite{DBLP:conf/eenergy/NeupaneSP17}. FOs are first generated by the prosumer agent from previous forecasts and analysis, and then sent to the aggregator, who decides whether to accept the FO or not and notifies its decision to the prosumer. If the FO is rejected, the process ends here; if it is accepted, the FO is aggregated to other FOs and optimized with them. After this, a schedule is determined for the FO and sent back to the prosumer agent for execution.

\subsection{Standard FlexOffers}
\label{subsection:SFO}

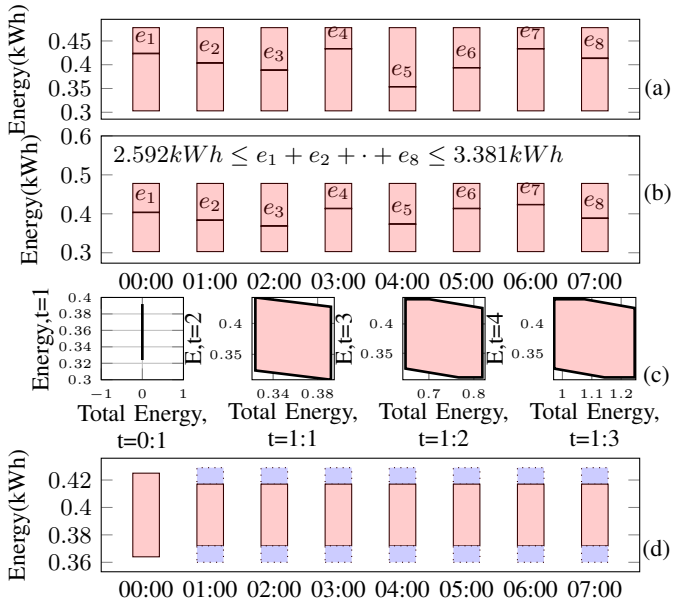
\begin{figure}
    \centering
%
%
\tikzstyle{every node}=[font=\small]
%
\begin{tikzpicture}

\begin{axis}[
width=0.48\textwidth,
height=0.16\textwidth,
ylabel=Energy(kWh),
legend style={at={(1,0.96)}},
xticklabels={},
xtick={1,2,3,4,5,6,7,8}, 
grid=major,
major grid style={draw=none},
ybar stacked,
xtick style={draw=none},
clip = false
]

\addplot[ybar, white,fill=white,opacity=0] coordinates {
    (1, 0.303) 
    (2, 0.303) 
    (3, 0.303) 
    (4, 0.303)
    (5, 0.303)
    (6, 0.303)
    (7, 0.303)
    (8, 0.303)
};

\addplot[ybar, red!20!black,fill=white!80!red] coordinates {
    (1, 0.12) 
    (2, 0.1) 
    (3, 0.085) 
    (4, 0.13)
    (5, 0.05)
    (6, 0.09)
    (7, 0.13)    
    (8, 0.11)    
};

\addplot[ybar, red!20!black, 
 nodes near coords,
 nodes near coords align=horizontal,
 nodes near coords style={
    anchor=south,
 },
 point meta=explicit symbolic
] coordinates {
    (1, 0.001) [$e_1$]
    (2, 0.001) [$e_2$]
    (3, 0.001) [$e_3$]
    (4, 0.001) [$e_4$]
    (5, 0.001) [$e_5$]
    (6, 0.001) [$e_6$]
    (7, 0.001) [$e_7$]    
    (8, 0.001) [$e_8$]   
};

\addplot[ybar, red!20!black,fill=white!80!red] coordinates {
    (1, 0.054) 
    (2, 0.074) 
    (3, 0.089) 
    (4, 0.044)
    (5, 0.124)
    (6, 0.084)
    (7, 0.044)    
    (8, 0.064)    
};

\node[] at (0.42\textwidth,0.35) {(a)};

\end{axis}

\begin{axis}[
width=0.48\textwidth,
height=0.18\textwidth,
at={(0, -0.105\textwidth)},
ymax=0.6,
ylabel=Energy(kWh),
xticklabels={{00:00}, {01:00}, {02:00}, {03:00}, {04:00}, {05:00}, {06:00}, {07:00}},
xtick={1,2,3,4,5,6,7,8}, 
grid=major,
major grid style={draw=none},
ybar stacked,
xtick style={draw=none},
clip = false
]

\addplot[ybar, white,fill=white,opacity=0] coordinates {
    (1, 0.303) 
    (2, 0.303) 
    (3, 0.303) 
    (4, 0.303)
    (5, 0.303)
    (6, 0.303)
    (7, 0.303)
    (8, 0.303)
};

\addplot[ybar, red!20!black,fill=white!80!red] coordinates {
    (1, 0.1) 
    (2, 0.08) 
    (3, 0.065) 
    (4, 0.11)
    (5, 0.07)
    (6, 0.11)
    (7, 0.12)    
    (8, 0.085)    
};

\addplot[ybar, red!20!black, 
 nodes near coords,
 nodes near coords align=horizontal,
 nodes near coords style={
    anchor=south,
 },
 point meta=explicit symbolic
] coordinates {
    (1, 0.001) [$e_1$]
    (2, 0.001) [$e_2$]
    (3, 0.001) [$e_3$]
    (4, 0.001) [$e_4$]
    (5, 0.001) [$e_5$]
    (6, 0.001) [$e_6$]
    (7, 0.001) [$e_7$]    
    (8, 0.001) [$e_8$]   
};

\addplot[ybar, red!20!black,fill=white!80!red] coordinates {
    (1, 0.074) 
    (2, 0.094) 
    (3, 0.109) 
    (4, 0.064)
    (5, 0.104)
    (6, 0.064)
    (7, 0.054)    
    (8, 0.089)    
};

\node[] at (4,0.55) {$2.592 kWh \leq e_1 + e_2 + \cdot + e_8 \leq 3.381  kWh$};

\node[] at (0.42\textwidth,0.45) {(b)};

\end{axis}

\begin{axis}[%
width=0.06\textwidth,
height=0.06\textwidth,
at={(0\textwidth,-0.19\textwidth)},
scale only axis,
xmin=-1,
xmax=1,
xtick={-1,0,1},
xlabel style={align=center},
xlabel={Total Energy,\\t=0:1},
xmajorgrids,
ymin=0.3,
ymax=0.4,
ylabel={Energy,t=1},
ymajorgrids,
axis background/.style={fill=white},
title style={font=\small},ylabel style={font=\small, yshift=-0.5em},xlabel style={font=\small, yshift=0.5em},ticklabel style={font=\tiny},legend style={font=\small, inner xsep=1pt, inner ysep=1pt,nodes={inner sep=1pt,text depth=0.05em}},
]

\addplot[area legend,solid,line width=1.0pt,draw=black,fill=white!80!red,forget plot]
table[row sep=crcr] {%
x	y\\
0	0.324\\
0	0.392\\
}--cycle;
\end{axis}

\begin{axis}[%
width=0.06\textwidth,
height=0.06\textwidth,
at={(0.11\textwidth,-0.19\textwidth)},
scale only axis,
xmin=0.321,
xmax=0.395,
xlabel style={align=center},
xlabel={Total Energy,\\t=1:1},
xmajorgrids,
ymin=0.309,
ymax=0.442,
ylabel={E,t=2},
ymajorgrids,
axis background/.style={fill=white},
xtick={0.34,0.38},
title style={font=\small},ylabel style={font=\small, yshift=-0.5em},xlabel style={font=\small, yshift=0.5em},ticklabel style={font=\tiny},legend style={font=\small, inner xsep=1pt, inner ysep=1pt,nodes={inner sep=1pt,text depth=0.05em}},
]

\addplot[area legend,solid,line width=1.0pt,draw=black,fill=white!80!red,forget plot]
table[row sep=crcr] {%
x	y\\
0.324	0.442\\
0.324	0.324\\
0.392	0.309\\
0.392	0.427\\
}--cycle;
\end{axis}

\begin{axis}[%
width=0.06\textwidth,
height=0.06\textwidth,
at={(0.22\textwidth,-0.19\textwidth)},
scale only axis,
xmin=0.642,
xmax=0.825,
xlabel style={align=center},
xlabel={Total Energy,\\t=1:2},
xmajorgrids,
ymin=0.305,
ymax=0.445,
ylabel={E,t=3},
ymajorgrids,
axis background/.style={fill=white},
xtick={0.7,0.8},
title style={font=\small},ylabel style={font=\small, yshift=-0.5em},xlabel style={font=\small, yshift=0.5em},ticklabel style={font=\tiny},legend style={font=\small, inner xsep=1pt, inner ysep=1pt,nodes={inner sep=1pt,text depth=0.05em}},
]

\addplot[area legend,solid,line width=1.0pt,draw=black,fill=white!80!red,forget plot]
table[row sep=crcr] {%
x	y\\
0.701   0.442\\
0.648	0.442\\
0.648	0.324\\
0.766   0.309\\
0.819	0.309\\
0.819	0.427\\
}--cycle;
\end{axis}

\begin{axis}[%
width=0.06\textwidth,
height=0.06\textwidth,
at={(0.33\textwidth,-0.19\textwidth)},
scale only axis,
xmin=0.97,
xmax=1.25,
xlabel style={align=center},
xlabel={Total Energy,\\t=1:3},
xmajorgrids,
ymin=0.305,
ymax=0.445,
ylabel={E,t=4},
ymajorgrids,
axis background/.style={fill=white},
title style={font=\small},ylabel style={font=\small, yshift=-0.5em},xlabel style={font=\small, yshift=0.5em},ticklabel style={font=\tiny},legend style={font=\small, inner xsep=1pt, inner ysep=1pt,nodes={inner sep=1pt,text depth=0.05em}},
]

\addplot[area legend,solid,line width=1.0pt,draw=black,fill=white!80!red,forget plot]
table[row sep=crcr] {%
x	y\\
1.075   0.442\\
0.972	0.442\\
0.972	0.324\\
1.143   0.309\\
1.246	0.309\\
1.246	0.427\\
}--cycle;

\end{axis}
\node[] at (0.405\textwidth,-3.4) {(c)};

\begin{axis}[
width=0.48\textwidth,
height=0.17\textwidth,
at={(0, -0.33\textwidth)},
ylabel=Energy(kWh),
xticklabels={{00:00}, {01:00}, {02:00}, {03:00}, {04:00}, {05:00}, {06:00}, {07:00}},
xtick={1,2,3,4,5,6,7,8}, 
grid=major,
major grid style={draw=none},
ybar stacked,
xtick style={draw=none},
clip = false
]

\addplot[ybar, white,fill=white,opacity=0] coordinates {
    (1, 0.364) 
    (2, 0.360) 
    (3, 0.360) 
    (4, 0.360)
    (5, 0.360)
    (6, 0.360)
    (7, 0.360)
    (8, 0.360)
};

\addplot[ybar, dotted, red!20!black,fill=blue!20!white] coordinates {
    (1, 0) 
    (2, 0.012) 
    (3, 0.012) 
    (4, 0.012)
    (5, 0.012)
    (6, 0.012)
    (7, 0.012)    
    (8, 0.012)    
};

\addplot[ybar, red!20!black,fill=white!80!red] coordinates {
    (1, 0.061) 
    (2, 0.045) 
    (3, 0.045) 
    (4, 0.045)
    (5, 0.045)
    (6, 0.045)
    (7, 0.045)    
    (8, 0.045)    
};

\addplot[ybar, dotted, red!20!black,fill=blue!20!white] coordinates {
    (1, 0) 
    (2, 0.012) 
    (3, 0.012) 
    (4, 0.012)
    (5, 0.012)
    (6, 0.012)
    (7, 0.012)    
    (8, 0.012)    
};

\end{axis}
\node[] at (0.405\textwidth,-5.7) {(d)};

\end{tikzpicture}%
    \caption{Example SFO (a), TECFO (b), DFO (c) and UFO (d).}
    \label{fig:SFO1}
\end{figure}

The simplest FO constraints are start time constraints and slice (energy) constraints. We are considering time as discrete and divided into time units, also called \emph{time slices}: their duration is one hour in the running example. Start time constraints define the earliest and latest time at which the load can start. Energy constraints define, at each time unit, the minimum and maximum amount of energy that can be consumed from the considered energy load: in other words, at each time unit $t$, two numbers $emin_t$ and $emax_t$ are defined such that $emin_t\leq e_t\leq emax_t$, where $e_t$ is the energy consumption during time unit $t$. An FO with just these constraints is called a \emph{Standard FlexOffer} (SFO). We can now give a formal definition of energy flexibility.
\begin{Definition}
With the notation just used, at each time unit $t$, we define \emph{amount flexibility} as $emax_t - emin_t$. Also, if the load can start between time units $t_0$ and $t_1$, we define \emph{time flexibility} as $t_1 - t_0$.
\end{Definition}
We can see an SFO example in Figure~\ref{fig:SFO1}(a) for the running example for eight time units. The pink bars indicate the amount of energy that can be consumed at each time unit: here, at each time unit, it varies between 0.303 kWh and 0.478 kWh. At each time, the horizontal bars indicate a possible amount of energy consumption $e_t$.

\subsection{Total Energy Constraint FlexOffers}

The total energy constraint (TEC) is a more complex type of constraint compared to SFO constraints: it defines a lower ($TE_{min}$) and an upper ($TE_{max}$) bound for the total consumption over all the considered time slices. This can be expressed as $TE_{min}\leq \sum\limits_{t=1}^T e_t \leq TE_{max}$.
An FO with slice and total energy constraints is called a \emph{Total Energy Constraint FO} (TECFO). Figure~\ref{fig:SFO1}(b) shows this for the running example: we have the slice constraints of the SFO example, plus a TEC defined by $TE_{min} = 2.592$ kWh and $TE_{max} = 3.381$ kWh.

\subsection{Dependency FlexOffers}

A dependency energy constraint defines dependencies between time slices. At each time unit $t$, a lower and an upper bound on the amount of consumable energy are defined, depending on how much energy has been consumed until time unit $t-1$. Formally, this can be expressed as $a \cdot (e_1+\ldots+e_{t-1} )+b\cdot e_t\leq c$, where $a,b,c$ are real numbers. An FO with dependency energy constraints is called a \emph{Dependency FlexOffer} (DFO). We can see an example in Figure~\ref{fig:SFO1}(c), for the first four time units of the running example: here, at each time unit, the $x$ axis represents the amount of energy consumed up to that time, while the $y$ axis represents the amount of consumable energy at that time. We can see that, in general, the more energy has been consumed up to time $t$, the less energy can be consumed at time $t$, and vice versa.
\subsection{Uncertain FlexOffers}
\label{section:UFO}

\emph{Uncertain FOs} (UFOs)~\cite{DBLP:conf/eenergy/LilliuPSN23} are a type of FOs that consider uncertainty from flexibility devices behavior. Two main types of uncertainty are considered: \emph{time} and \emph{amount} uncertainty. At each time unit, time uncertainty refers to the probability for the device to be operative at that time, while amount uncertainty refers to the amount of energy that can be consumed by the device at that time.
A UFO is created in two steps. First, uncertainty related to the state (in the running example, temperature) is modeled at each time $t$; second, the probability for each energy value at each time to be feasible is calculated, taking into account both types of uncertainty. Some functions $\{f_1,\ldots,f_T\}$ describing those probabilities will be obtained: those functions will define the UFO.
UFOs can be visualized by choosing a probability threshold $P_0$. At each time $t$, the energy values having probability at least $P_0$ of being available for consumption can be described by intervals. Figure~\ref{fig:SFO1}(d) shows this visualization for the running example: if we choose $P_0 = 1$, at each time unit, the available energy values are described by the pink bars, while if we choose $P_0 = 0.8$, the available energy values are described by the combined pink and blue bars. This is because at the first time unit, the status of the device is known, and the amount of available flexibility can be determined exactly, while for the next time units, there is more uncertainty (represented here by the blue bars) and the amount of flexibility certainly available (pink bars) is lower. 

\subsection{SAREF and SAREF4ENER}
\label{subsection:SAREF4ENER}

\emph{Smart Applications REFerence} (SAREF)\footnote{https://saref.etsi.org/}~\cite{SAREF} is an ontology for IoT devices and solutions published by ETSI\footnote{\url{https://www.etsi.org/}} in a series of Technical Specifications initially released in 2015 and updated in 2017. Even though its initial objective was to build a reference ontology for appliances relevant to energy efficiency, SAREF can also serve as a reference model to enable better integration of data from various vertical domains in the IoT. SAREF has been extended to different domains
: specifically, \emph{SAREF for Energy Flexibility} (SAREF4ENER)~\cite{SAREF4ENER} is its extension for the energy domain, and aims to provide a standardized way to describe and interoperate energy-related devices, systems, and services.
It allows us to describe devices such as solar panels, electric vehicles, and smart meters, as well as systems like microgrids, and energy management/demand response services. 
By using SAREF4ENER, developers can create applications that easily integrate and communicate with different energy devices, systems, and services, regardless of the manufacturer or protocol used, which enables more efficient and sustainable energy management.

\subsection{dCO}
\begin{figure*}[!ht]
    \centering
    \includegraphics[width=0.8\linewidth]{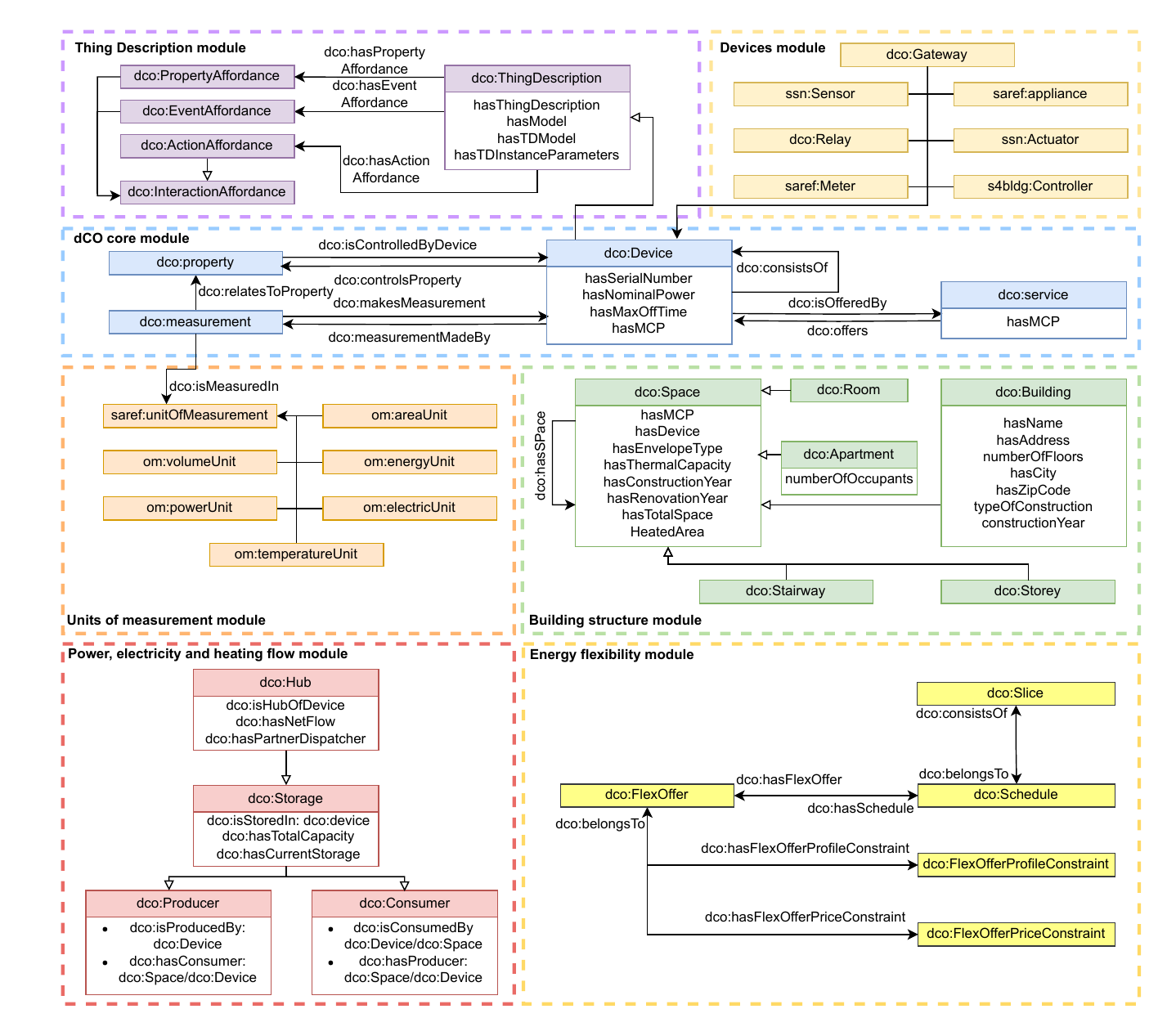}
    \caption{dCO design overview.}
    \label{fig:dcooverview}
\end{figure*}
The \emph{World Wide Web Consortium} (W3C) has developed a standard for enabling Internet of Things (IoT) interoperability, whose name is \emph{W3C Web of Things} (WoT)~\cite{WoT}. In this context, \emph{Thing Descriptions} (TDs)~\cite{WoT_TD} have been developed: they describe the metadata and interfaces of the considered devices that participate in the WoT. Things are organized in TD Directories (WoT TDDs), which make it possible to register, manage, and search TDs. An important feature for IoT consumers is to send queries to the WoT directory: this is achieved by having a common ontology to annotate TDs; however, before dCO, no ontologies were integrating WoT into smart buildings. dCO~\cite{DBLP:conf/esws/LaadharTP22} has been created with the purpose of achieving this integration. Specifically, dCO is an ontology designed for smart buildings, capable of representing devices and properties of the building, in addition to energy flexibility: what makes it different from other building ontologies is the integration with TDs. In this paper, we will focus on the aspect of dCO that strictly concerns energy flexibility. 
We now quickly describe the main modules. 
The \texttt{core} module uses the SAREF core ontology~\cite{SAREF}, and enables the integration of the other modules in order to allow interoperability: it is integrated with the other modules. The \texttt{TD} module integrates the WoT TD ontology~\cite{TDontology} with the \texttt{core} module, while the \texttt{devices module} represents the devices. The \texttt{units of measurement} module use the Units of Measurement Ontology~\cite{UOMontology}, while the \texttt{building description} module represents building-related metadata and building topology, and uses the W3C Building Topology Ontology~\cite{BOTontology}. All these modules have been described in~\cite{DBLP:conf/esws/LaadharTP22} in greater detail. Finally, there are the \texttt{power, electricity and heating flow} module, which is self-explanatory, and the \texttt{energy flexibility} module: this last is the contribution of this work, where we explain how it extends SAREF4ENER. Figure~\ref{fig:dcooverview} shows those modules, and their main classes and properties.

\section{dCO Extension of SAREF4ENER}
\label{section:extension}

This section shows how SAREF4ENER has been extended to describe FOs. We show how we designed the dCO flexibility module and how we encode attributes and constraints. 

\subsection{FlexOffer Message}
\label{section:FOmessage1}
As mentioned in Section~\ref{section:preliminaries}, FOs are sent between actors: this is achieved by FO messages, which encode FO metadata and information about energy constraints.
The attributes of an FO message describe the FO ID, state (e.g., accepted, rejected), time granularity, geographical location, time limits for acceptance and assignment, constraints on energy, time and price, and an energy schedule. Of these, the only ones to have an equivalent in SAREF4ENER are state, simple energy constraints (slice, TECs), price constraints and duration, while more advanced energy constraints (for DFOs and UFOs) and other attributes had to be added as extensions. 

\subsection{Energy Flexibility Ontological Design}
\label{section:design}
\begin{figure*}
    \centering
    \includegraphics[scale=.8]{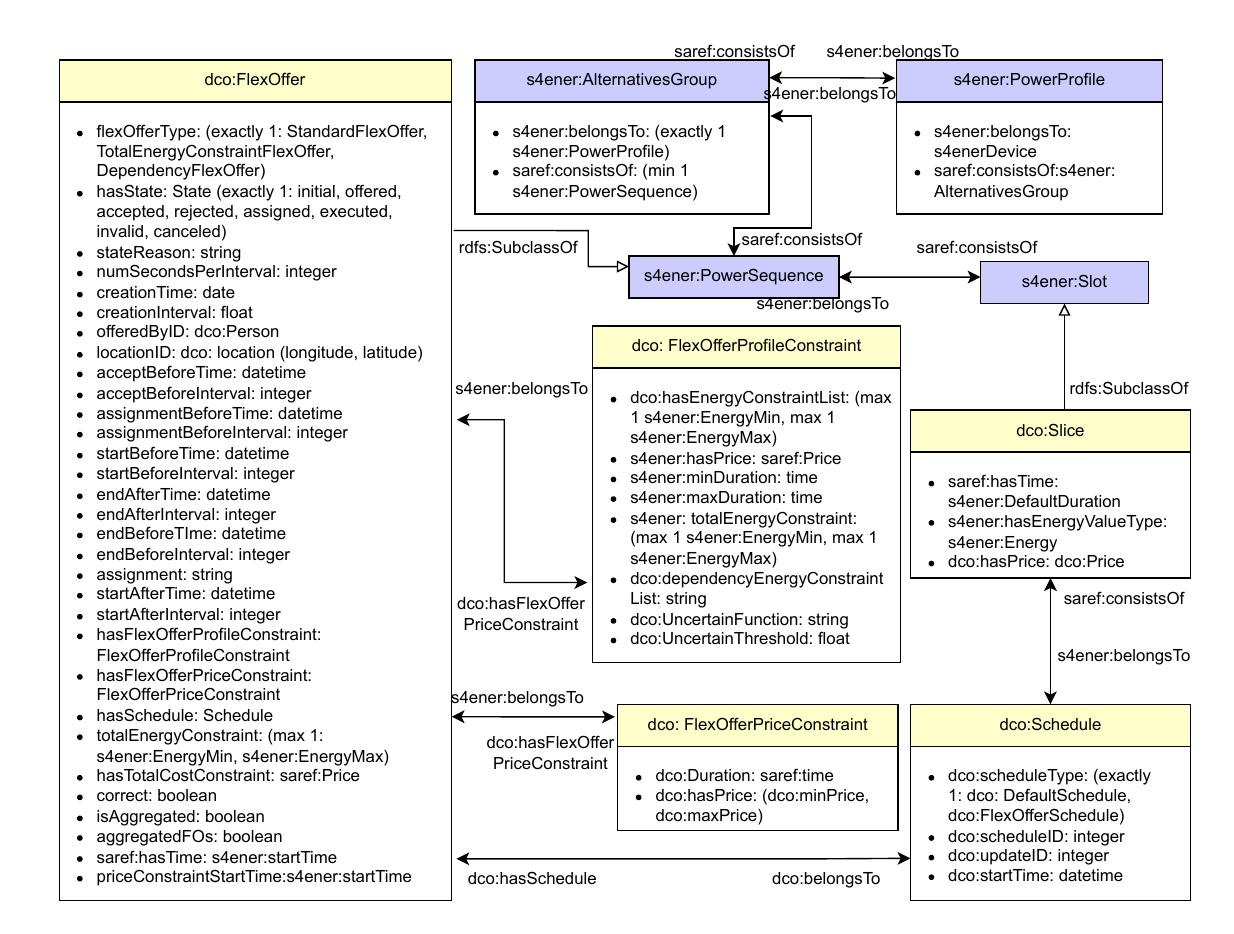}
    \caption{Overview of the flexibility module of dCO.}
    \label{fig:dCO}
\end{figure*}

We now describe how dCO has been designed for flexibility. A complete overview can be seen in Figure~\ref{fig:dCO}.
In the figure, we highlighted in purple the existing SAREF4ENER (written with prefix \texttt{s4ener}) classes that we used, and in yellow the new dCO (written with prefix \texttt{dco}) that we have defined. In this section, we use the notation \emph{O:e} to describe the element \emph{e} belonging to the ontology \emph{O}. The core class for FO representation is \emph{dco:FlexOffer}: it is defined as a subclass of \emph{s4ener:PowerSequence}, the SAREF4ENER class describing a power sequence, meaning that \emph{dco:FlexOffer} inherits the object properties and data properties of \emph{s4ener:PowerSequence}. Specifically, \emph{dco:FlexOffer} consists of at least one power sequence slot, represented by \emph{s4ener:Slot}. Regarding the other dCO classes, \emph{dco:FlexOffer} consists of at least one profile constraint, denoted by the class \emph{dco:FlexOfferProfileConstraint}. This class has the following properties:
\begin{itemize}
    \item \emph{dco:hasEnergyConstraintList}: (max 1 \emph{s4ener:EnergyMin}, max 1 \emph{s4ener:EnergyMax}). In Figure~\ref{fig:SFO1}(a), they are 0.303 and 0.478 respectively);
\item \emph{s4ene:hasPrice}: sarefPrice (contains the list(s) of price constraints for one-time unit, in EUR);
\item \emph{s4ener:MinDuration}: The minimum duration of a FlexOffer, in time units; in Figure~\ref{fig:SFO1}(a), it is 8;
\item \emph{s4ener:MaxDuration}: The maximum duration of a FlexOffer, in time units; in Figure~\ref{fig:SFO1}(a), it is 8;
\item \emph{dco:totalEnergyConstraint}: (max 1 \emph{s4ener:EnergyMin}, max 1 \emph{s4ener:EnergyMax}). In Figure~\ref{fig:SFO1}(b), they are 2.592 and 3.381 respectively);
\item \emph{dco:dependencyEnergyConstraintList}: Contains the dependency constraints for energy in a Dependency FlexOffer;
\item \emph{dco:uncertainFunctions}: (min 1 Function): Contains the parameters which define the uncertain FlexOffer;
\item \emph{dco:uncertainThreshold}: a real number that represents the probability threshold. In the running example in Figure~\ref{fig:SFO1}(d), it can be 1 (pink) or 0.8 (blue).
\end{itemize}

Also, \emph{dco:FlexOffer} may have one or more price constraints, denoted by the class \emph{dco:FlexOfferPriceConstraint}, which has the following properties: \emph{dco:Duration}: \emph{saref:Time}; \emph{dco:hasPrice}: \emph{saref:Price}.

Finally, \emph{dco:FlexOffer} can have one or more schedules, represented by \emph{dco:Schedule}, having the following properties:
\begin{itemize}
    \item \emph{dco:scheduleType}: can be one of the two classes: \emph{dco:DefaultSchedule} or \emph{dco:FlexOfferSchedule};
    \item \emph{dco:scheduleID}: the id of a schedule;
\item \emph{dco:updateID}: the id of the schedule update;
\item \emph{dco:startTime}: the start time of a schedule.
\end{itemize}
Regarding \emph{dco:scheduleType}, there are two choices because FOs may have two schedules, i.e., a default schedule (\emph{dco:defaultSchedule}), which indicates a default baseload, or an FO schedule (\emph{dco: FlexOfferSchedule}), which indicates an FO-optimized schedule. 
Also, \emph{dco:Schedule} consists of at least one slice, represented by \emph{dco:Slice}, which is a subclass of \emph{s4ener:Slot}. Specifically, \emph{dco:Slice} inherits all properties of \emph{s4ener:Slot}, and extends it with the \emph{dco:hasPrice} object property, which defines the energy price of a slice. In the running example, the SFO shown in Figure~\ref{fig:SFO1}(a) has eight slices. 
A \emph{dco:slice} is associated with the following:
\begin{itemize}
    \item \emph{s4ener:DefaultDuration}: the duration of a \emph{dco:Slice};
    \item \emph{s4ener:hasEnergyValueType}: range is either \emph{s4ener: Energy} or \emph{s4ener:Power}. Here, we are considering energy;
    \item \emph{s4ener:hasPrice}: the unitary price (in EUR/kWh) of energy in a \emph{dco:Slice}.
\end{itemize}

We finally show the properties of the \emph{dco:FlexOffer} class:
\begin{itemize}
\item \emph{dco:hasState}: State of the FO (initial/offered/ accepted/rejected/assigned/executed/ invalid/ canceled);
\item \emph{dco:stateReason}: Reason for FO rejection (if State == rejected);
\item \emph{dco:locationID}: the location of the FO in the grid system;
\item \emph{dco:acceptBeforeInterval}: Interval before which FO must be accepted;
\item \emph{dco:assignmentBeforeInterval}: Interval before which FO must be scheduled;
\item \emph{dco:startBeforeInterval} / \emph{dco:startAfterInterval}: Interval before / after which FO must be started;
\item \emph{dco:endBeforeInterval} / \emph{dco:endAfterInterval}: Interval before / after which FO execution must end;
\item \emph{dco:acceptanceBeforeTime}: Absolute time before which FO must be accepted;
\item \emph{dco:assignmentBeforeTime}: Absolute time before which FO must be scheduled;
\item \emph{dco:numSecondsPerInterval}: Duration (in seconds) of a time interval. The default value is 900, but in the running example, we consider it as 3600;
\item \emph{dco:startBeforeTime} / \emph{dco:startAfterTime}: Absolute time before / after which FO must be started (in the running example, 00:00 of the considered day);
\item \emph{dco:endBeforeTime} / \emph{dco:endAfterTime}: Absolute time before / after which FO execution must end (in the running example, 08:00 of the considered day);
\item \emph{dco:creationTime}: Absolute time at which the FO has been created;
\item \emph{dco:priceConstraintStartTime}: The start time of the price constraint.
\end{itemize}


\begin{table*}[h!]
    \centering
 \begin{tabular}{|l|c|c|l|l|}
    \hline
\textbf{Element} & \makecell[c]{\textbf{Mandatory} \\ \textbf{in the FO?}} & \makecell[c]{\textbf{Exists in SAREF/}\\\textbf{SAREF4ENER?}} & \makecell[c]{\textbf{Corresponding class in }\\ \textbf{SAREF/SAREF4ENER}} & \textbf{Notes}\\
\hline
\multicolumn{5}{|c|}{Standard FlexOffer}\\
\hline
ID & Yes & No &  & \\
\hline
State & Yes & Yes & \href{https://saref.etsi.org/saref4ener/PowerSequenceState}{PowerSequenceState} & \\
\hline
StateReason & No & No & & 	\\
\hline
NumSecondsPerInterval & Yes & No &  & 		\\
\hline
CreationTime & Yes & No &  & \\
\hline
CreationInterval & Yes & No &  & 		\\
\hline
OfferedByID & Yes & No &  & \\
\hline
LocationID & No & No &  & In SAREF, a device can have a location \\
\hline
AcceptBeforeTime & No & No & & \\
\hline
AcceptBeforeInterval & No & No &  & \\
\hline
AssignmentBeforeTime & No & No &  & \\
\hline
AssignmentBeforeInterval & No & No &  &  \\
\hline
StartAfterInterval & No & No &  & 	\\
\hline
StartBeforeTime & Yes & No &  & \\
\hline
StartBeforInterval & No & No &  & \\
\hline
EndAfterTime & No & No &  & \\
\hline
EndAfterInterval & No & No &  & \\
\hline
EndBeforeTime & No & No &  & \\
\hline
EndBeforeInterval & No & No &  & \\
\hline
StartAfterTime & Yes & No &  & \\
\hline
FlexOfferProfileConstraints & Yes & No &  & \\
\hline
FlexOfferPriceConstraint & No & No &  & \\
\hline
DefaultSchedule & No & No &  & \\
\hline
FlexOfferSchedule & No & No &  & \\
\hline
\multicolumn{5}{|c|}{FlexOfferProfileConstraints}\\
\hline
EnergyConstraintsList & Yes & Yes & \href{https://saref.etsi.org/saref4ener/Energy}{Energy}  & \makecell{ 
SAREF includes \href{https://saref.etsi.org/saref4ener/EnergyMax}{EnergyMax}
\\and 
\href{https://saref.etsi.org/saref4ener/EnergyMin}{EnergyMin}}\\
\hline
PriceConstraint & No & Partially & \href{https://saref.etsi.org/core/hasPrice}{hasPrice}   & 
\makecell{SAREF does not include min \\and max price}\\
\hline
MinDuration & No & Yes & \href{https://saref.etsi.org/saref4ener/ActiveDurationMin}{ActiveDurationMin} & \\
\hline
MaxDuration & No & Yes & \href{https://saref.etsi.org/saref4ener/ActiveDurationMax}{ActiveDurationMax}  & \\
\hline
TotalCostConstraint & No & No &  & \\
\hline
\multicolumn{5}{|c|}{ScheduleSlice attributes:}\\
\hline
Duration & No & Yes & \href{https://saref.etsi.org/saref4ener/v1.1.2/#s4ener:DefaultDuration}{DefaultDuration}  & \\
\hline
EnergyAmount & Yes & Yes & \href{https://saref.etsi.org/saref4ener/EnergyExpected}{EnergyExpected} & \\
\hline
Price & No & No &  & \\
\hline
 \end{tabular}
    \caption{FlexOffer mapping to SAREF4ENER - core message.}
    \label{tab:SAREF}
\end{table*}

\begin{table*}[h!]
    \centering
    \begin{tabular}{|l|c|c|l|l|}
    \hline
\textbf{Element} & \makecell[c]{\textbf{Mandatory} \\ \textbf{in the FO?}} & \makecell[c]{\textbf{Exists in SAREF/}\\\textbf{SAREF4ENER?}} & \makecell[c]{\textbf{corresponding class in }\\ \textbf{SAREF/SAREF4ENER}} & \textbf{Notes}\\
\hline
\multicolumn{5}{|c|}{Total energy FlexOffer}\\
\hline
\makecell[l]{TotalEnergyConstraint\\(lower and upper)} & No & Yes & \href{https://saref.etsi.org/saref4ener/Energy}{Energy} & \makecell{SAREF includes \href{https://saref.etsi.org/saref4ener/EnergyMax}{EnergyMax}
\\and 
\href{https://saref.etsi.org/saref4ener/EnergyMin}{EnergyMin}}\\
\hline
\multicolumn{5}{|c|}{Dependency FlexOffer}\\
\hline
DependencyEnergyConstraintList
 & No & No &  & \\
\hline
\makecell[l]{PriceConstraint\\(minPrice, maxPrice)} & No & Partially & \href{https://saref.etsi.org/core/hasPrice}{hasPrice}  & \\
\hline
MinDuration & No & Yes & \href{https://saref.etsi.org/saref4ener/ActiveDurationMin}{ActiveDurationMin} & \\
\hline
MaxDuration & No & Yes & \href{https://saref.etsi.org/saref4ener/ActiveDurationMax}{ActiveDurationMax}  & \\
\hline
\multicolumn{5}{|c|}{Uncertain FlexOffer}\\
\hline
UncertainFunctions & No & No &  & \\
\hline
UncertainThreshold & No & No &  & \\
\hline
MinDuration & No & Yes & \href{https://saref.etsi.org/saref4ener/ActiveDurationMin}{ActiveDurationMin} & \\
\hline
MaxDuration & No & Yes & \href{https://saref.etsi.org/saref4ener/ActiveDurationMax}{ActiveDurationMax}  & \\
\hline
 \end{tabular}
    \caption{FlexOffer mapping to SAREF4ENER.}
    \label{tab:SAREF2}
\end{table*}

\subsection{Constraints Encoding}
\label{section:encoding}

Here, we show how we encode the FO constraints into the dCO ontology: for those constraints, we will use the same notation we used in Section~\ref{section:preliminaries}. As we will see in Section~\ref{section:comparison}, the only properties in SAREF4ENER that can be used for this are \emph{s4ener:EnergyMin} and \emph{s4ener:EnergyMax}: it is possible to represent SFOs and TECFOs by mapping to them $emin_t$ and $emax_t$ for slice constraints, and $TE_{min}$ and $TE_{max}$ for TECs. 

Regarding DFOs, we show how we encoded the constraints. Consider the running example, and the DFO represented in Figure~\ref{fig:SFO1}(c). The slices are represented by polygons, which are intersections of hyperplanes: thus, we can write these polygons as sets of inequalities over $x$ and $y$. We can then build a matrix, with as many rows as the number of inequalities defining the polygon: in this case, the number of rows would be six. As the inequalities are linear, they can be written as $ax + by \leq c$: the corresponding row of the matrix will then be $[a\;b\;c]$. This is called \emph{H-representation} of the DFO~\cite{DBLP:conf/eenergy/SiksnysP16}. We show below the constraints and matrix representation of the fourth slice. The property \emph{DependencyEnergyConstraintList} encodes the matrix.

\begin{minipage}{4.2cm}
    \begin{equation*}
    \begin{aligned}
        y &\geq 0.309\\
        y &\leq 0.442\\
        x & \geq 0.972\\
        x & \leq 1.246\\
        0.088\cdot x + y &\geq 0.41\\
        0.088\cdot x + y &\leq 0.537
    \end{aligned}
\end{equation*}
\end{minipage}
\begin{minipage}{4.2cm}
\begin{equation*}
    \begin{bmatrix}
0 & -1 & -0.309\\
0 & 1 & 0.442\\
-1 & 0 & -0.972\\
1 & 0 & 1.246\\
-0.088 & -1 & -0.41\\
0.088 & 1 & 0.537
\end{bmatrix}
\end{equation*}
\end{minipage}

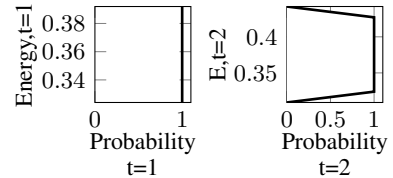
\begin{figure}
    \centering
%
%
\tikzstyle{every node}=[font=\small]
%
\begin{tikzpicture}
\begin{axis}[%
width=0.07\textwidth,
height=0.07\textwidth,
at={(0\textwidth,-0.46\textwidth)},
scale only axis,
xmin=0,
xmax=1.1,
xtick={-1,0,1},
xlabel style={align=center},
xlabel={Probability\\t=1},
ymin=0.324,
ymax=0.392,
ylabel={Energy,t=1},
axis background/.style={fill=white},
title style={font=\small},ylabel style={font=\small, yshift=-0.5em},xlabel style={font=\small, yshift=0.5em},ticklabel style={font=\small},legend style={font=\small, inner xsep=1pt, inner ysep=1pt,nodes={inner sep=1pt,text depth=0.05em}},
]

\addplot[area legend,solid,line width=1.0pt,draw=black,fill=white!80!red,forget plot]
table[row sep=crcr] {%
x	y\\
1	0.324\\
1	0.392\\
}--cycle;
\end{axis}

\begin{axis}[%
width=0.07\textwidth,
height=0.07\textwidth,
at={(0.14\textwidth,-0.46\textwidth)},
scale only axis,
xmin=0,
xmax=1.1,
xlabel style={align=center},
xlabel={Probability\\t=2},
ymin=0.309,
ymax=0.442,
ylabel={E,t=2},
axis background/.style={fill=white},
title style={font=\small},ylabel style={font=\small, yshift=-0.5em},xlabel style={font=\small, yshift=0.5em},ticklabel style={font=\small},legend style={font=\small, inner xsep=1pt, inner ysep=1pt,nodes={inner sep=1pt,text depth=0.05em}},
]

\addplot[area legend,solid,line width=1.0pt,draw=black,fill=white,forget plot]
table[row sep=crcr] {%
x	y\\
0	0.309\\
1	0.324\\
1	0.427\\
0	0.442\\
};
\end{axis}

\end{tikzpicture}%
    \caption{UFO functions}
    \label{fig:UFO}
\end{figure}

Finally, UFOs are described by the functions $\{f_1, \ldots, f_T\}$ mentioned in Section~\ref{section:UFO}. Describing a generic function is nontrivial; however, for the devices supported by FOs, we noticed that they can be written as $f_t = \min\{p_1,\ldots,p_{N_t}\}$, where each $p_i$ is a polynomial defined in the range of possible energy values. In the running example, a representation of $f_1$ and $f_2$ can be seen in Figure~\ref{fig:UFO}: those functions are, however, represented on the $y$ axis rather than the $x$ axis, because FOs represent energy over the $y$ axis. We can see that for $t = 1$ the function is the constant polynomial $f_1 = 1$, while for $t = 2$ we have $f_2 = \min\{p_1,p_2,p_3\}$, where $p_1 = 1$, $p_2 = 66.67 \cdot x - 20.6$, and $p_3 = -66.67 \cdot x + 29.467$. Once the polynomials have been determined, they are encoded as a list of lists of coefficients, with the terms corresponding to each degree in increasing order. For example, since $f_1 = 1$, the UFO constraint for $t = 1$ is encoded as $[1]$. On the other hand, for $t = 2$, the list will be $[[1],[-20.6,66.67],[29.467,-66.67]]$. 

\subsection{FO and SAREF4ENER attribute comparison}
\label{section:comparison}

In this section we map all the FO message attributes into SAREF4ENER, whenever possible, to observe which FO attributes are present in SAREF4ENER and which ones are missing. Table~\ref{tab:SAREF} and Table~\ref{tab:SAREF2} show this mapping, describing, for each attribute of the FO message, whether it is mandatory or not, whether it exists in SAREF or SAREF4ENER or not, what is the corresponding class in SAREF/SAREF4ENER, and notes for additional information. 
More precisely, Table~\ref{tab:SAREF2} shows the mapping for the main FO message attributes and slice and time constraints, while Table~\ref{tab:SAREF2} shows the mapping for the more general constraints, such as TEC, dependency, and uncertainty.
We can see that, among the core FO message attributes, only the attribute \emph{State} has a direct counterpart in SAREF4ENER, \emph{PowerSequenceState}, while other attributes are not directly supported. Duration and slice constraints can be mapped to SAREF4ENER, while SAREF has a partial match for \emph{PriceConstraint}, although it does not allow specification of minimum and maximum prices. FO schedules can also be mapped to SAREF4ENER, except for price attributes, and the same applies to TECs.
However, dependency and uncertainty constraints cannot be mapped to SAREF4ENER, as shown in the lower part of the table. Consequently, SAREF4ENER is not capable of representing DFOs and UFOs, and thus  cannot benefit from the accurate flexibility representation given by these advanced types of FOs.

\section{Evaluation}
\label{section:evaluation}

This section shows the evaluation of this ontology. Specifically, how it has been demonstrated in real life and integrates with the current standards, how it has been validated, and how it compares with the current state of the art.

\subsection{Real-life demonstration and standard integration}

The EU project in which dCO was developed demonstrated its feasibility through real-world case studies conducted in Denmark, France, and Switzerland. All functionalities, e.g., querying building descriptions, are operational, and interoperability between smart buildings and services has been successfully achieved. dCO supports the flexibility services developed in this project. 

The Swiss demonstrator site manages the flexibility of a pool of buildings, with the goal of contributing to the energy market. All buildings are equipped with heat pumps and are grouped into a single pool to aggregate their flexibility potential. Flexibility is realized via load shedding: FOs are generated with the possibility of temporarily turning off heat pumps and, if the optimization process indicates this action and comfort constraints are respected, the selected heat pumps are switched off. The site includes 63 buildings, and the service has been active for 18 months, from July 2022 to December 2023. The average daily flexibility for the entire pool varies significantly, being less than 40 kWh/day during summer and having its highest values in winter, reaching above 255 kWh/day during January.


The French demonstrator site provides a digital energy coaching service for flexibility. A dedicated application notifies grid users in case of a potential power outage, prompting them to reduce or shift their loads in exchange for a financial incentive. This service has been deployed to 124 individual customers over a 50-day period, arbitrarily selected between March and July 2023, yielding several MWh of flexibility.



The Danish demonstrator site provides services of room-based heating control and cost optimization, by using flexibility provided by heat pumps. This service was deployed for 11 buildings throughout 2023. Results show a 13\% cost reduction compared to the baseline, and an average reduction of nearly 20\% in energy demand during the three most expensive hours of the day.

For all those demonstrators, dCO contributed making the development of these services easier and better. There are two main reasons for this: i) services can automatically identify the available devices they refer to, as dCO annotates TDs, and are compatible with multiple platforms; ii) dCO supports FOs, which allow for better flexibility representation compared to simple flexibility representations and for better scalability compared to exact flexibility models, as we will see in Section~\ref{section:resultscompare}. 

Regarding the integration of this ontology with existing standards or protocols, our ontology allows for better flexibility modeling, being able to capture more complex energy constraints and integrate FOs. Our proposed ontology is also compatible with the Common Information Model (CIM)\footnote{\url{https://www.entsoe.eu/digital/common-information-model/}}, which is the standard from the International Electrotechnical Commission for electric power transmission and distribution, and is capable of representing more complex energy constraints than CIM, which is unable to represent dependency and uncertain energy constraints~\cite{CIM}. Finally, SFOs and TECFOs integrate with EEBus\footnote{\url{https://www.eebus.org/}}, another energy data standard that only has simple flexibility models: this is because EEBus builds on SAREF4ENER, which supports slice and total energy constraints. 


\subsection{Validation}

In order to validate dCO, we also used the \emph{OntOlogy Pitfall Scanner!} (OOPS!) tool\footnote{\url{https://oops.linkeddata.es/}} to evaluate it. OOPS!~\cite{poveda2014oops} is a tool, well-known in the Semantic Web community, designed with the purpose of identifying pitfalls and design errors in ontologies. There are three levels of importance in pitfalls identified by OOPS!. \textbf{Critical} pitfalls need to be resolved, as they may affect the ontology consistency, reasoning, and applicability. \textbf{Important} pitfalls would improve the ontology's robustness if fixed, but they do not undermine the ontology's function. \textbf{Minor} pitfalls would improve the ontology's readability if fixed, but do not pose problems for the ontology.
OOPS! identified quite a few minor and important pitfalls, but only 11 critical ones. However, the critical pitfalls were all of the type \emph{Defining multiple domains or ranges in properties}. In our case, this was not only valid, but also \emph{necessary} for properties to have multiple domains and ranges to capture the complexity of our flexibility scenario. A simple example of this is the \emph{dco:hasEnergyConstraintList} property, which can have up to one \emph{s4ener:EnergyMin} attribute and up to one \emph{s4ener:EnergyMax} attribute. For this reason, we \emph{intentionally} designed our ontology with this feature, and therefore none of the critical pitfalls identified by the OOPS! tool undermines the function of the ontology, rather, they are exactly as they should be. Thus, the ontology meets the checked requirements. 

\subsection{Comparison with SAREF4ENER}
\label{section:resultscompare}

In Section~\ref{subsection:SAREF4ENER}, we mentioned that SAREF4ENER is an ontology that allows for a simple description of energy flexibility. FOs, on the other hand, is a model that represents flexibility in mathematical terms; despite having two different purposes, a parallel can be drawn between SAREF4ENER and FO representations, in that they refer to energy profiles and time slices. However, this parallel highlights a crucial weakness of SAREF4ENER in describing energy flexibility. Specifically, 
SAREF4ENER is unable to model many necessary attributes for FOs: in the main FO message, only \emph{State} 
and 
the attributes defining slice and total energy constraints can be mapped to SAREF4ENER. It is therefore \emph{possible} to map SFO and TECFO constraints into SAREF4ENER. However, as mentioned in Section~\ref{section:FOmessage1}, dependency constraints and UFOs \emph{cannot} be mapped to SAREF4ENER. On the other hand, the dCO flexibility module has been built in order to provide full support for all types of FOs. This means not only that dCO can represent flexibility through more advanced constraints, but also that it allows for communication via a FO message. This way, by using dCO, developers have the possibility to design applications able to exploit the whole range of benefits coming from FO modeling, i.e., i) advanced and more accurate flexibility representation, ii) support for more advanced devices, and iii) really fast optimization and (dis)aggregation of flexibility.

Regarding flexibility representation, SAREF4ENER is able at most to accurately represent flexibility for devices that only need this type of constraint. 
This is the case with wet devices, such as dishwashers and washing machines, as shown in~\cite{DBLP:conf/smartgridcomm/PedersenSN18}. These devices have a load with a fixed power/amount profile, which is only shifted in time. 
However, for other devices, SFOs/TECFOs are outclassed by DFOs in terms of accuracy in flexibility representation, as we can see in previous literature. The ability to retain flexibility for a model is calculated by the \emph{economic metric}~\cite{Lilliu2021}, defined as the amount of profit that can be obtained by optimizing flexibility with the objective of cost reduction. Specifically, for the charge and discharge process of batteries, DFOs retain $61$\% of the flexibility, against the $38$\% from TECFOs and $10$\% from SFOs~\cite{Lilliu2021}. UFOs can perform even better, retaining up to $66.4$\% of flexibility, which goes up to $86.8$\% of flexibility when imbalance penalties for violating constraints are high (\emph{high penalty} case)~\cite{DBLP:conf/eenergy/LilliuPSN23}. Regarding EVs, flexibility modeling via DFOs is more accurate than SFOs and TECFOs, as slice constraints and TECs are only a subset of DFO constraints, and therefore the amount of flexibility retained by DFOs is higher: yet, in~\cite{DBLP:conf/eenergy/LilliuPSN23}, we see that UFOs can retain $92$\% of flexibility, against the $77.3$\% retained by DFOs. For the high penalty case, UFOs retain $87.5$\% of flexibility, against the $61.8$\% from DFOs, which still outperform SFOs and TECFOs. Finally, regarding heat pumps, the comparison between DFOs and SFOs has been done in~\cite{DBLP:conf/eenergy/SiksnysP16}, showing that DFOs can retain more than three times the amount of flexibility from SFOs: in~\cite{DBLP:conf/eenergy/LilliuPS23} DFOs for heat pumps have been further improved, showing that they can retain up to $98.9$\% of flexibility according to the economic metric. We show these results in Table~\ref{tab:results}. The ability for dCO to model the significant amounts of flexibility for these devices ensures that developers can create applications that allow for an accurate description of flexibility for batteries, EVs and heat pumps. Table~\ref{tab:FOsvsS4E} summarizes this, showing which use cases are supported by the dCO flexibility module and SAREF4ENER, respectively.

\begin{table}[h!]
    \centering
    \begin{tabular}{|l|c|c|c|c|}
    \hline
Device & SFO & TECFO & DFO & UFO\\
\hline
Battery & 10\% & 38\% & 61\% & \makecell{66.4\%\\(86.8\%)} \\
\hline
EV & N/A & N/A & \makecell{77.3\%\\(61.8\%)} & \makecell{92\%\\(87.5\%)}  \\
\hline
Heat pumps & N/A & N/A & 98.9\% & N/A \\
\hline
    \end{tabular}
    \caption{Amount of retained flexibility (in parenthesis, the high penalty case).}
    \label{tab:results}
\end{table}

\begin{table}[h!]
    \centering
    \begin{tabular}{|l|c|c|}
    \hline
Use case & dCO & SAREF4ENER\\
\hline
Operating wet devices & \checkmark & \checkmark\\
\hline
Charging and discharging batteries & \checkmark & \\
\hline
Charging and discharging EVs & \checkmark & \\
\hline
Operating heat pumps & \checkmark & \\
\hline
    \end{tabular}
    \caption{Use cases supported.}
    \label{tab:FOsvsS4E}
\end{table}

Furthermore, FOs are extended over time to represent more use cases and flexibility properties: an example is their capability of modeling different energy vectors from electricity, such as heat~\cite{DBLP:conf/eenergy/LilliuPS23}. This has been our rationale to extend SAREF4ENER to represent both FOs in general and, more specifically, DFO and UFO constraints: the resulting ontology can represent all the advanced use cases presented in this paper, as well as provide a solid foundation for future ontology extensions based on new types of FOs.  

Therefore, we have shown two important facts. First, DFOs and UFOs vastly outperform SFOs and TECFOs in terms of accuracy of flexibility representation. As it is possible to map only SFOs and TECFOs to SAREF4ENER, the flexibility representation allowed by dCO is much more accurate than the flexibility representation allowed by SAREF4ENER. Second, as shown in Table~\ref{tab:FOsvsS4E}, dCO allows for accurately representing many more types of devices compared to SAREF4ENER: specifically, batteries, EVs and heat pumps can be accurately represented by dCO, but not by SAREF4ENER.

Lastly, dCO allows for support for the fast optimization, aggregation and disaggregation of energy loads provided by FOs. Specifically, FOs allow to aggregate, optimize and disaggregate up to $2 \cdot 10^6$ of loads for 96 time units time horizons~\cite{DBLP:conf/eenergy/LilliuPS23}. In comparison, an exact approach described in~\cite{DBLP:conf/eenergy/LilliuPS23} can only optimize up to 500 devices for that time horizon, while other known exact approaches cited in~\cite{DBLP:conf/eenergy/LilliuPS23} fail already to perform optimization for more than 6 time units.

\section{Conclusions and Future Work}
\label{section:conclusion}

Energy flexibility models are capable of accurate representations of flexibility, and FOs in particular can do that while being scalable for long time horizons and many devices. For their real-life application, it is important to allow flexibility in data exchange with different types of devices in several contexts, like smart buildings. Standardization of data formats is therefore required, and ontologies fulfill this task. Ontologies have evolved over the years, and SAREF4ENER is the current industry standard ontology for describing the energy usage of smart devices. However, SAREF4ENER fails to support flexibility for relevant use cases such as batteries, EVs, and heat pumps. This paper aims to cover this gap by proposing an extension of SAREF4ENER capable of modeling FOs: we integrated this extension into the dCO ontology, as its flexibility module. This extension allows for the representation of FO message attributes and, specifically, FO constraints: this way, advanced constraints such as dependency and uncertainty constraints can be encoded, and it is possible to accurately represent nontrivial use cases such as the ones mentioned above. 
Future work will be centered on extending the dCO ontology with newly developed FO features for handling more use cases and flexibility properties. Regarding dCO maintenance, it will follow the evolution of FOs, which evolve but are always backwards-compatible by design. Therefore, maintenance of the dCO flexibility module will only need to encode eventual new energy constraints, while the core part of the message will not need significant changes.

\bibliographystyle{IEEEtran}
\bibliography{bibltrue}

\appendix


\subsection{FlexOffer Message}
\label{section:FOmessageandusecase}
Here, we describe FO messages and their attributes.
Table~\ref{tab:FOattributes} shows a full description of an FO message, outlining which attributes are present, whether they are mandatory or not, their data type, and a short description of each attribute. Examples of the message can be seen in Appendix~\ref{section:sfoappendix} onwards: the core part of the message is described first, while the part of the message that describes the constraints changes depending on which constraints are used, and is shown in Appendix~\ref{section:sfoappendix} through~\ref{section:UFOappendix}. 

\begin{table*}[h!]
    \centering
    \begin{tabular}{|l|c|l|l|}
    \hline
Attribute &	Mandatory &	Type &	Description\\
\hline
ID & Yes & Integer & The ID that identifies the FO\\
\hline
State & Yes & String & \makecell[l]{State of the FO\\ (initial/offered/accepted/rejected/assigned/executed/ invalid/ canceled)}\\
\hline
StateReason & No & String & Reason for FO rejection (if State == rejected)\\
\hline
NumSecondsPerInterval & Yes & Integer & Duration (in seconds) of a time interval. The default value is 900\\
\hline
CreationTime & Yes & Datetime & Absolute time at which the FO has been created\\
\hline
CreationInterval & No & Integer & \makecell[l]{FO creation interval calculated as epoch value for the \\CreationTime / NumSecondsPerInterval ratio}\\
\hline
OfferedByID & Yes & String & ID of the FO owner\\
\hline
LocationID & No & String & ID for representing the location of the FO in the grid system\\
\hline
AcceptBeforeTime & No & Datetime & Absolute time before which FO must be accepted\\
\hline
AcceptBeforeInterval & No & Integer & Interval before which FO must be accepted\\
\hline
AssignmentBeforeTime & No & Datetime & Absolute time before which FO must be scheduled\\
\hline
AssignmentBeforeInterval & No & Integer & Interval before which FO must be scheduled\\
\hline
StartAfterTime & Yes & Datetime & Absolute time after which FO must be started\\
\hline
StartAfterInterval & No & Integer & Interval after which FO must be started\\
\hline
StartBeforeTime & Yes & Datetime & Absolute time before which FO must be started\\
\hline
StartBeforeInterval & No & Integer & Interval before which FO must be started\\
\hline
EndAfterTime & No & Datetime & Absolute time after which FO execution must end\\
\hline
EndAfterInterval & No & Integer & Interval after which FO execution must end\\
\hline
EndBeforeTime & No & Datetime & Absolute time before which FO execution must end\\
\hline
EndBeforeInterval & No & Integer & Interval before which FO execution must end\\
\hline
FlexOfferProfileConstraints & Yes & FlexOfferSlice & Constraints for FO profile\\
\hline
FlexOfferPriceConstraint & No & PriceSlice & Constraints for FO price\\
\hline
DefaultSchedule & No & ScheduleSlice & Default energy consumption and time schedule of the FO\\
\hline
FlexOfferSchedule & No & ScheduleSlice & Schedule for the FO\\
\hline
\multicolumn{4}{|c|}{FlexOfferProfileConstraints attributes}\\
\hline
EnergyConstraintsList & Yes & Object & \makecell[l]{Contains the list(s) of energy constraints for one time unit, expressed \\in kWh. Has two sub-elements: lower, and upper.}\\
\hline
PriceConstraint & No & Object & \makecell[l]{Contains the list(s) of price constraints for one time unit. \\Has two sub-elements : lower, and upper.}\\
\hline
MinDuration & No & Integer & Minimum duration of the FO\\
\hline
MaxDuration & No & Integer & Maximum duration of the FO\\
\hline
TotalEnergyConstraint & No & Float & List of total energy constraints. Has two sub-elements : lower, and upper.\\
\hline
TotalCostConstraint & No & Float & List of total cost constraints.\\
\hline
\multicolumn{4}{|c|}{ScheduleSlices attributes}\\
\hline
Duration & No & Integer & Indicates the duration of the considered slices, in time units.\\
\hline
EnergyAmount & Yes & Float & Indicates the energy consumption for those slices.\\
\hline
Price & No & Float & Indicates the price amount for that slice.\\
\hline
    \end{tabular}
    \caption{FlexOffer message attributes.}
    \label{tab:FOattributes}
\end{table*}


The attributes shown in Table~\ref{tab:FOattributes} compose a message: in the following, we show an example for that.

\begin{lstlisting}[language=json,numbers=none]
"flexOffer": {
    "id": "4188a132-a937-4639-96cf-d8529fa78b86",
    "state": "Adaptation",
    "stateReason": "FlexOffer Initialized",
    "creationInterval": 1726911,
    "offeredById": "harry@80060B5E0FD671D58243CE7162A6054719822955",
    "locationId": {
        "userLocation": {"longitude": 9.990595, "latitude": 57.012293}
    },
    "acceptanceBeforeInterval": 1726914,
    "assignmentBeforeInterval": 1726914,
    "startAfterInterval": 1726912,
    "startBeforeInterval": 1726920,
    "assignment": "obligatory",
    "flexOfferProfileConstraints": {$\ldots$}
    "acceptanceBeforeTime": "2019-04-02T16:30:00.000+0000",
    "assignmentBeforeTime": "2019-04-02T16:30:00.000+0000",
    "numSecondsPerInterval": 900,
    "startAfterTime": "2019-04-02T16:00:00.000+0000",
    "startBeforeTime": "2019-04-02T18:00:00.000+0000",
    "creationTime": "2019-04-02T15:45:00.000+0000",
    "correct": true
\end{lstlisting}

Here, we omitted the element FlexOfferProfileConstraints, which describes the constraints defining the FO: its sub-elements are described in the second half of Table~\ref{tab:FOattributes}.

\subsection{SFO}
\label{section:sfoappendix}
Below we can see an example of how the FlexOfferProfileConstraints attribute would appear in the case of the SFO produced for the running example, that we can see in Figure~\ref{fig:SFO1}(a). In the case we are showing there are also price constraints, where the minimum price for the considered time units is 0.03€, and the maximum is 0.15€.
\begin{lstlisting}[language=json,numbers=none]
"flexOfferProfileConstraints": [{        
            "energyConstraintList": [{"lower": 0.303, "upper": 0.478}],
            "priceConstraint": {"minPrice": 0.03, "maxPrice": 0.15},
            "minDuration": 1,
            "maxDuration": 1
        },{
            "energyConstraintList": [{"lower": 0.303, "upper": 0.478}],
            "priceConstraint": {"minPrice": 0.03, "maxPrice": 0.15},
            "minDuration": 1,
            "maxDuration": 1
        },{
            "energyConstraintList": [{"lower": 0.303, "upper": 0.478}],
            "priceConstraint": {"minPrice": 0.03, "maxPrice": 0.15},
            "minDuration": 1,
            "maxDuration": 1
        },{
            "energyConstraintList": [{"lower": 0.303, "upper": 0.478}],
            "priceConstraint": {"minPrice": 0.03, "maxPrice": 0.15},
            "minDuration": 1,
            "maxDuration": 1
        },{
            "energyConstraintList": [{"lower": 0.303, "upper": 0.478}],
            "priceConstraint": {"minPrice": 0.03, "maxPrice": 0.15},
            "minDuration": 1,
            "maxDuration": 1
        },{
            "energyConstraintList": [{"lower": 0.303, "upper": 0.478}],
            "priceConstraint": {"minPrice": 0.03, "maxPrice": 0.15},
            "minDuration": 1,
            "maxDuration": 1
        },{
            "energyConstraintList": [{"lower": 0.303, "upper": 0.478}],
            "priceConstraint": {"minPrice": 0.03, "maxPrice": 0.15},
            "minDuration": 1,
            "maxDuration": 1
        },{
            "energyConstraintList": [{"lower": 0.303, "upper": 0.478}],
            "priceConstraint": {"minPrice": 0.03, "maxPrice": 0.15},
            "minDuration": 1,
            "maxDuration": 1
        }]
\end{lstlisting}

Data in ScheduleSlice format has the attributes described in Table~\ref{tab:FOattributes}, in the bottom part.

We can also see that Figure~\ref{fig:SFO1}(a) contains a schedule. This is how it would be encoded in an FO message, assuming this is the default schedule:
\begin{lstlisting}[language=json,numbers=none]
"defaultSchedule": {
        "scheduleId": 0,
        "updateId": 0,
        "scheduleSlices": [{"duration": 1, "energyAmount": 0.423, "price": 0.05},
{"duration": 1, "energyAmount": 0.403, "price": 0.1},
{"duration": 1, "energyAmount": 0.388, "price": 0.1},
{"duration": 1, "energyAmount": 0.433, "price": 0.03},
{"duration": 1, "energyAmount": 0.353, "price": 0.03},
{"duration": 1, "energyAmount": 0.393, "price": 0.05},
{"duration": 1, "energyAmount": 0.433, "price": 0.07},
{"duration": 1, "energyAmount": 0.413, "price": 0.07}],
        "startTime": "2019-04-02T00:00:00.000+0000"
    }
\end{lstlisting}
\subsection{TECFO}
For TECFOs the FO message looks the same as for SFOs, except there is an additional TEC at the end of it. Specifically, consider the TECFO shown in Figure~\ref{fig:SFO1}(b), relative to the running example. The \emph{flexOfferProfileConstraints} attribute would appear like this:

\begin{lstlisting}[language=json,numbers=none]
"flexOfferProfileConstraints": [{        
            "energyConstraintList": [{"lower": 0.303, "upper": 0.478}],
            "priceConstraint": {"minPrice": 0.03, "maxPrice": 0.15},
            "minDuration": 1,
            "maxDuration": 1
        },{
            "energyConstraintList": [{"lower": 0.303, "upper": 0.478}],
            "priceConstraint": {"minPrice": 0.03, "maxPrice": 0.15},
            "minDuration": 1,
            "maxDuration": 1
        },{
            "energyConstraintList": [{"lower": 0.303, "upper": 0.478}],
            "priceConstraint": {"minPrice": 0.03, "maxPrice": 0.15},
            "minDuration": 1,
            "maxDuration": 1
        },{
            "energyConstraintList": [{"lower": 0.303, "upper": 0.478}],
            "priceConstraint": {"minPrice": 0.03, "maxPrice": 0.15},
            "minDuration": 1,
            "maxDuration": 1
        },{
            "energyConstraintList": [{"lower": 0.303, "upper": 0.478}],
            "priceConstraint": {"minPrice": 0.03, "maxPrice": 0.15},
            "minDuration": 1,
            "maxDuration": 1
        },{
            "energyConstraintList": [{"lower": 0.303, "upper": 0.478}],
            "priceConstraint": {"minPrice": 0.03, "maxPrice": 0.15},
            "minDuration": 1,
            "maxDuration": 1
        },{
            "energyConstraintList": [{"lower": 0.303, "upper": 0.478}],
            "priceConstraint": {"minPrice": 0.03, "maxPrice": 0.15},
            "minDuration": 1,
            "maxDuration": 1
        },{
            "energyConstraintList": [{"lower": 0.303, "upper": 0.478}],
            "priceConstraint": {"minPrice": 0.03, "maxPrice": 0.15},
            "minDuration": 1,
            "maxDuration": 1
        },
            {"TotalEnergyConstraints": [{"lower": 2.592, "upper": 3.381}]}
    ]
\end{lstlisting}
with the slice constraints listed for each time unit, and the total energy constraint at the end.

\subsection{DFO}
\label{section:DFOappendix}

The flexOfferProfileConstraints attribute for DFOs looks different from SFOs and TECFOs, as it utilizes the H-representation of the DFO and therefore encodes matrices, instead of lower and upper bounds. An example for the first four time units of the running example, shown in Figure~\ref{fig:SFO1}(c), can be seen below:

\begin{lstlisting}[language=json,numbers=none]
"flexOfferProfileConstraints": [{        
            "DependencyEnergy ConstraintList": [[0, 1, 0.392],[0, -1, -0.324]],
            "priceConstraint": {"minPrice": 0.03, "maxPrice": 0.15},
            "minDuration": 1,
            "maxDuration": 1
        },{
            "DependencyEnergy ConstraintList": [[-1, 0, -0.324],[1, 0, 0.392],[-0.221, -1, -0.396],[0.221, 1, 0.514]],
            "priceConstraint": {"minPrice": 0.03, "maxPrice": 0.15},
            "minDuration": 1,
            "maxDuration": 1
        },{
            "DependencyEnergy ConstraintList": [[0, -1, -0.309],[0, 1, 0.442],[-1, 0, -0.648],[1, 0, 0.819],[-0.127, -1, -0.406],[0.127, 1, 0.531]],
            "priceConstraint": {"minPrice": 0.03, "maxPrice": 0.15},
            "minDuration": 1,
            "maxDuration": 1
        },{
            "DependencyEnergy ConstraintList": [[0, -1, -0.309],[0, 1, 0.442],[-1, 0, -0.972],[1, 0, 1.246],[-0.088, -1, -0.41],[0.088, 1, 0.537]],
            "priceConstraint": {"minPrice": 0.03, "maxPrice": 0.15},
            "minDuration": 1,
            "maxDuration": 1
        },
    ]
\end{lstlisting}

At each time, the DependencyEnergyConstraintList is the representation of a matrix, represented by its rows. Specifically, the one at $t = 4$ is the same matrix we described in Section~\ref{section:encoding}.

\subsection{UFO}
\label{section:UFOappendix}

Finally, UFO constraints are different from other FO constraints, as they are functions. However, they can be represented by sets of polynomials, as shown in Section~\ref{section:encoding}: each polynomial is represented by its coefficients, written in increasing order of degree. We can see here the representation of the UFO shown in Figure~\ref{fig:UFO}:

\begin{lstlisting}[language=json,numbers=none]
"flexOfferProfileConstraints": [{        
            "UncertainEnergy ConstraintList": [[1]],
            "priceConstraint": {"minPrice": 0.03, "maxPrice": 0.15},
            "minDuration": 1,
            "maxDuration": 1
        },{
            "DependencyEnergy ConstraintList": [[1],[-20.6,66.67],[29.467,-66.67]],
            "priceConstraint": {"minPrice": 0.03, "maxPrice": 0.15},
            "minDuration": 1,
            "maxDuration": 1
        }
    ],

\end{lstlisting}

\end{document}